\documentclass[pdflatex,sn-nature]{sn-jnl}

\usepackage{graphicx}%
\usepackage{multirow}%
\usepackage{amsmath,amssymb,amsfonts}%
\usepackage{amsthm}%
\usepackage{mathrsfs}%
\usepackage[title]{appendix}%
\usepackage{xcolor}%
\usepackage{textcomp}%
\usepackage{manyfoot}%
\usepackage{booktabs}%
\usepackage{algorithm}%
\usepackage{algorithmicx}%
\usepackage{algpseudocode}%
\usepackage{dashrule}
\usepackage{listings}%
\usepackage{subcaption}
\usepackage{tabularx}
\usepackage[section]{placeins}
\usepackage{xr}
\usepackage{caption}

\theoremstyle{thmstyleone}%
\theoremstyle{thmstyletwo}%

\theoremstyle{thmstylethree}%

\begin{document}

\title[Article Title]{Information-Theoretic Multi-Model Fusion for Target-Oriented Adaptive Sampling in Materials Design}

\author*[1]{\fnm{Yixuan} \sur{Zhang}}\email{yixuan.zhang@tmm.tu-darmstadt.de}

\author[1]{\fnm{Zhiyuan} \sur{Li}}\email{zhiyuan.li@tmm.tu-darmstadt.de}

\author[2]{\fnm{Weijia} \sur{He}}\email{weijia.he@stud.tu-darmsatdt.de}

\author[1]{\fnm{Mian} \sur{Dai}}\email{mian.dai@tu-darmstadt.de}

\author[2]{\fnm{Chen} \sur{Shen}}\email{chenshen@tmm.tu-darmstadt.de}

\author*[2]{\fnm{Teng} \sur{Long}}\email{tenglong@sdu.edu.cn}

\author*[1]{\fnm{Hongbin} \sur{Zhang}}\email{hzhang@tmm.tu-darmstadt.de}

\affil[1]{\orgdiv{Institute of Materials Science}, \orgname{Technical University of Darmstadt}, \orgaddress{\street{Otto-Berndt-Straße 3}, \city{Darmstadt}, \postcode{64287}, \country{Germany}}}

\affil[2]{\orgdiv{School of Materials Science and Engineering}, \orgname{Shandong University}, \orgaddress{\street{Jingshi Road 17923}, \city{Jinan}, \postcode{250061}, \state{Shandong}, \country{China}}}



\abstract{Target-oriented discovery under limited evaluation budgets requires making reliable progress in high-dimensional, heterogeneous design spaces where each new measurement is costly, whether experimental or high-fidelity simulation. We present an information-theoretic framework for target-oriented adaptive sampling that reframes optimization as trajectory discovery: instead of approximating the full response surface, the method maintains and refines a low-entropy information state that concentrates search on target-relevant directions. The approach couples data, model beliefs, and physics/structure priors through dimension-aware information budgeting, adaptive bootstrapped distillation over a heterogeneous surrogate reservoir, and structure-aware candidate manifold analysis with Kalman-inspired multi-model fusion to balance consensus-driven exploitation and disagreement-driven exploration. Evaluated under a single unified protocol without dataset-specific tuning, the framework improves sample efficiency and reliability across 14 single- and multi-objective materials design tasks spanning candidate pools from $600$ to $4 \times 10^6$ and feature dimensions from $10$ to $10^3$, typically reaching top-performing regions within 100 evaluations. Complementary 20-dimensional synthetic benchmarks (Ackley, Rastrigin, Schwefel) further demonstrate robustness to rugged and multimodal landscapes.}

\keywords{Information Theory, Bayesian Optimization, Multi-Model Ensemble, Data-Scarce Materials Design}

\maketitle

\section{Introduction}\label{sec1}

Designing materials with optimal properties often entails navigating vast, high-dimensional parameter spaces where both experimental measurements and theoretical simulations are costly and limited in number\cite{lookman_information_2016,rajan_materials_2015, lookman_active_2019, agrawal_perspective_2016,wang2023scientific}. This data scarcity is particularly acute when multiple properties must be optimized simultaneously, and when the design variables span hundreds or thousands of features\cite{ren_accelerated_2018,zhang_strategy_2018,zunger_inverse_2018}. Classically, scientific discovery arises from the interplay between data, statistical inference, and physical modeling\cite{box_science_1976,mackay_information_2003}. However, when data are scarce and reliable physical models are unavailable, this triadic understanding collapses into purely statistical empiricism. While this shift aligns with the pragmatic reality that target acquisition (possession) precedes explanatory modeling (comprehension), it forces algorithms to navigate in a maximum-entropy environment without structural constraints, significantly increasing the information required to locat valid solutions~\cite{ramprasad_machine_2017,lookman_information_2016}. The resulting tension between limited data and high information demand defines a fundamental bottleneck in modern materials design\cite{karniadakis_physics-informed_2021,garnett2023bayesian}. To mitigate this gap, one may exploit latent physical information implicitly encoded in the structural organization of candidate pools and in the behavior of surrogate models, as both can carry prior judgments rooted in chemical intuition, thermodynamic feasibility, or empirical design heuristics\cite{ghiringhelli_big_2015,ward_general-purpose_2016,deringer_gaussian_2021} (cf. Figure~\ref{fig:intro}. a, upper half). Crucially, recent theoretical insights suggest that natural data and valid physical solutions typically reside on low-dimensional manifolds embedded within the high-dimensional ambient space~\cite{li2025basicsletdenoisinggenerative}. Consequently, the complexity of any target-relevant trajectory is often far lower than that of the full search domain\cite{gomez2018automatic}. The key challenge, therefore, is to shift the paradigm from exhaustively modeling the entire space to trajectory discovery, by establishing a coherent, low-entropy cognitive state that progressively converges along high-probability paths pointing toward the design goal\cite{shahriari_taking_2016, balachandran_adaptive_2016, tian_materials_2025}.

In computational terms, this setting can be viewed as target-oriented black-box optimization under extreme sampling constraints: the algorithm must make reliable progress with only tens of evaluations in an ambient space that may have hundreds to thousands of degrees of freedom. The central difficulty is not merely to approximate the global response surface, but to quickly identify and follow a small subset of target-relevant directions from sparse, noisy, and heterogeneous information sources, i.e., a low-entropy trajectory. This motivates a shift from exhaustive function modeling to trajectory discovery, where the goal is to construct a coherent information state that concentrates search effort onto high-probability paths leading to the design target.

From an information-theoretic perspective, realizing this shift requires constructing a tripartite cognitive system that dynamically aligns data, model beliefs, and physical constraints to maximize the mutual information with the design target\cite{qian2023knowledge, hennig2012entropy, hernandez2014predictive} (cf. Figure~\ref{fig:intro}a). Rather than treating these elements in isolation, the system must evolve as a coherent whole: by steering exploration toward regions that maximally reduce system entropy, the effective complexity of the problem decreases as the search progresses, projecting the intractable high-dimensional search onto a target-relevant manifold\cite{tenenbaum2000global, long2024generative}. Achieving rapid convergence under such constraints necessitates two complementary strategies: (i) Dimension-aware capacity alignment and information distillation, which restricts the search complexity to the intrinsic dimension of the data, thereby preventing models from overfitting to high-dimensional noise in irrelevant regions\cite{eriksson2019scalable, moriconi2020high} (cf. Figure~\ref{fig:intro}. b-d); and (ii) Structural information filtering and decision making, which leverages heterogeneous information channels to separate macroscopic trends from local fluctuations, extracting complementary signals that a single model would miss\cite{kennedy2000predicting, abuawwad2025kalman} (cf. Figure~\ref{fig:intro}. e-g).

\begin{center} 
    \includegraphics[width=\textwidth]{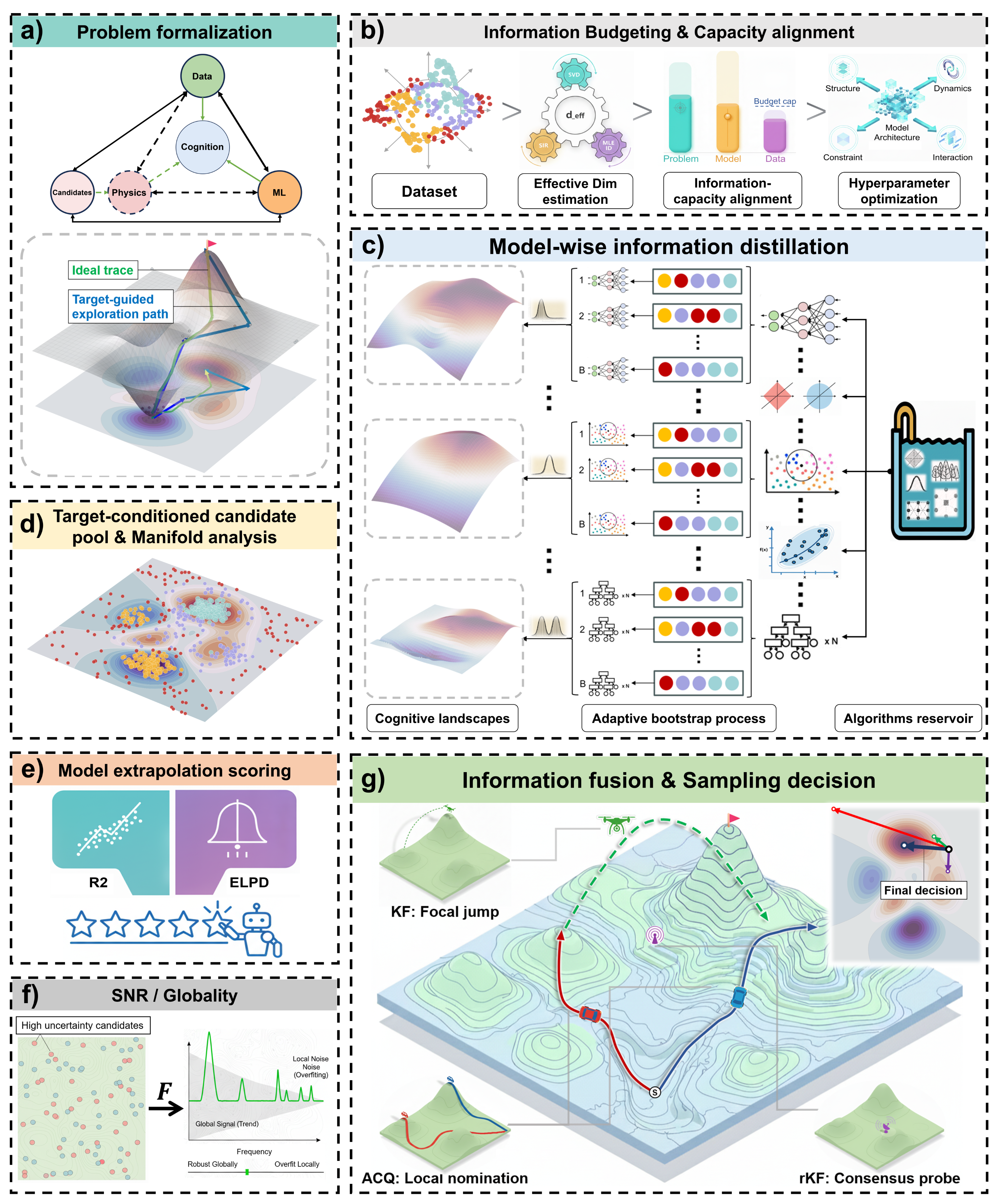}
\end{center}
\captionof{figure}{\textbf{Information-theoretic multi-source fusion pipeline for target-oriented adaptive sampling}. 
        \textbf{(a) Problem formalization}: We frame data, model beliefs, and physical constraints as a tripartite cognitive system that co-evolves to maximize mutual information with the design target, shifting the goal from exhaustive global modeling to discovering target-relevant trajectories.
        \textbf{(b) Information budgeting \& capacity alignment}: the effective information dimension implied by the current data is estimated and combined with the remaining evaluation budget to match model capacity to problem complexity, enabling hyperparameter adaptation that preserves target-relevant structure while filtering high-dimensional noise. 
        \textbf{(c) Model-wise information distillation}: a heterogeneous reservoir of models is trained under an adaptive bootstrap procedure, producing per-model “cognitive landscapes” that encode complementary inductive biases and uncertainty views. 
        \textbf{(d) Target-conditioned candidate generation and manifold analysis}: when a candidate set is provided, it is directly analyzed for its manifold structure; otherwise, target-relevant candidates are sampled from the collection of cognitive landscapes and then subjected to manifold analysis to characterize the feasible/structured region for search. 
        \textbf{(e) Model extrapolation scoring}: ut-of-bag evaluation during bootstrapping yields model generalization diagnostics, including point-prediction accuracy and distributional predictive quality 
        \textbf{(f) SNR and globality assessment}: combining model outputs with the candidate manifold, we quantify signal-to-noise characteristics and diagnose whether each model’s cognition is predominantly global or local.
        \textbf{(g) Information fusion and sampling decision making}: multiple fusion modes are used to translate heterogeneous signals into actions. Standard acquisition functions (ACQ) nominate locally promising regions, Kalman-inspired fusion (KF) aggregates multi-model evidence to enable consensus-driven focal jumps, and a uncertainty-weighted variant (rKF) prioritizes consensus probes that most efficiently reduce disagreement and accelerate convergence toward the target region.}
\label{fig:intro}

Traditional Gaussian process (GP)-based Bayesian optimization (BO), despite extensions like deep kernel learning, remains fundamentally mismatched to this high-dimensional, data-scarce regime\cite{wilson2016deep,ober2021promises}. A primary failure mode lies in the lack of dimension-aware capacity control: in high-dimensional ambient spaces, distance concentration degrades kernel validity, while data scarcity renders hyperparameters weakly identifiable\cite{aggarwal2001surprising,binois2022survey}. Without explicit constraints on the effective dimension, standard surrogates are prone to hallucination—overfitting to aliased high-frequency noise rather than capturing the underlying physics. Furthermore, relying on any single surrogate family imposes a rigid structural prior and a fixed spectral bias on the search\cite{rahaman2019spectral,fanaskov2023spectral}. A single model typically acts as a fixed-bandwidth filter, struggling to simultaneously capture the macroscopic trends (low-frequency signals) required for global coverage and the local fine-grained transitions (high-frequency signals) required for precision\cite{han2022learning,wu2021autoformer}. These limitations motivate a departure from single-model BO toward an adaptive framework that functions as a multi-scale inference system. Crucially, realizing the potential of this system requires more than passive aggregation; it demands an active information fusion mechanism to dynamically adjudicate between conflicting belief states\cite{lakshminarayanan2017simple,erickson2020autogluon}. By rigorously arbitrating between model consensus (exploitation) and informative disagreement (exploration) based on real-time reliability estimates, such a system can robustly separate physical structure from high-dimensional noise, ensuring that the search trajectory converges along the path of maximum information gain even when the underlying function landscape is poorly understood\cite{liang2018investigating,beluch2018power}.

Building on these principles, we introduce an information-theoretic trajectory discovery framework that redefines optimization as the entropy reduction of a tripartite cognitive system comprising data, model, and physics axes. Unlike traditional methods that treat these elements in isolation, our approach aligns them through a four-stage pipeline designed to function as a heterogeneous cognitive filter against high-dimensional noise. First, we enforce dimension-aware capacity control, dynamically estimating the effective intrinsic dimension to cap model complexity, ensuring surrogates capture learnable manifolds rather than overfitting to sparse data\cite{campadelli2015intrinsic, floryan2022data}. Second, we employ target-conditioned surrogate shaping, where heterogeneous models (linear, tree-based, neural) are trained via importance sampling on high-value regions\cite{kumar2020conservative, maraval2022sample}. This ensemble acts as a complementary bank where low-capacity models secure macroscopic trend coverage while high-capacity models resolve local nuances. Third, we introduce a structure-aware candidate organization layer that groups candidates and estimates structural weights using neighborhood and partition based analyses, thereby quantifying redundancy, local density, and feasible variation without requiring an explicit embedding or continuous-to-graph projection, safeguarding against mode collapse\cite{belkin2003laplacian, merchant2023scaling, tian2024boundary}. Finally, decision-making is driven by a multi-source fusion mechanism that leverages a Kalman-like logic to arbitrate between model consensus (exploitation) and high-value disagreement (exploration)\cite{maddox2019simple, ma2020kalman}. By explicitly maximizing the mutual information between this tripartite system and the target design, the framework constructs a robust search trajectory that remains effective even when individual models fail to capture the full physics\cite{wang2017max, moss2021gibbon}.

Applied to 14 single- and multi-objective materials design tasks spanning datasets of 600–4,000,000 samples, feature dimensions of 10–1025, and one to four target properties, the framework operates under a uniform, data-scarce setting: optimization starts from only 20 randomly selected poor initial samples whose target values smaller than 50\% of the global optimum, proceeds in batches of 10 new measurements per iteration, and typically converges within fewer than 10 iterations (100 evaluations). Despite this minimal sampling budget, the method consistently identifies top-10 candidates across all materials problems. Under the same configuration, it also performs robustly on 20-dimensional benchmark mathematical functions: reaching the global optimum for Ackley within 100 iterations, achieving ~50\% global-optimum convergence on Schwefel with the remainder near-optimal, and locating low-error local optima for Rastrigin with performance gaps below 50. Overall, the framework delivers robust performance gains across diverse problem classes under a unified, data-scarce protocol, without relying on problem-specific acquisition engineering or extensive hyperparameter tuning. By integrating physical priors from candidates, model diversity, and uncertainty into a single information-centric control loop, it supports adaptive sampling decisions that remain effective from synthetic high-dimensional functions to million-scale materials candidate pools.

\section{Results and Discussion}\label{sec2}

\subsection{Performance on mathematical benchmark functions}

We adopt three canonical 20D benchmarks: Ackley, Rastrigin, and Schwefel, because they expose complementary challenges that are prototypical for high-dimensional black-box optimization. Ackley features a nearly flat outer region surrounding a narrow central valley; in high dimensions the weak gradients outside the valley, compounded by distance concentration, provide little directional information, making the key difficulty to rapidly locate the informative valley and then contract along it rather than wasting samples on the quasi-plateau. Rastrigin exhibits dense periodic multimodality with large-amplitude local minima; barrier crossings often yield limited information gain because neighboring basins are similar, so global progress is easily confounded by strong local structure, testing the ability to form reliable global trends under extreme multimodality. Schwefel places the global optimum far from the origin amid many attractive local minima; naive trajectories are drawn into deceptive basins, and success requires occasional, well-informed long moves to reach the distant global basin. From a unifying perspective, if local optima are viewed as background ``noise'' and the global trend toward the true optimum as the ``signal,'' these benchmarks primarily differ in their signal-to-background contrast: Ackley is a localization problem under weak gradients, Rastrigin a disambiguation problem under high-amplitude local noise, and Schwefel a barrier-crossing problem with deceptive attractors. The progressively reduced contrast between global signal and local background across these functions provides a structured validation of the framework’s three essential capabilities: (i) the ability to actively gather and concentrate information to rapidly locate informative trajectories under severe data scarcity (Ackley); (ii) the ability to integrate heterogeneous information channels to disentangle local noise from global trends (Rastrigin); and (iii) the ability to trigger targeted, high-gain transitions that promote barrier crossing in deceptive landscapes (Schwefel).

All three benchmark functions were tested under a unified configuration to ensure comparability across landscapes of distinct complexity. Each function was evaluated in 20 dimensions within its conventional search domain: $[-32.768,\,32.768]^{20}$ for Ackley, $[-5.12,\,5.12]^{20}$ for Rastrigin, and $[-500,\,500]^{20}$ for Schwefel. To align with the framework’s default maximization formulation, each function was sign-inverted so that minimizing the original objective corresponds to maximizing its negative form. Optimization began from ten randomly selected low-quality initial points and proceeded in batches of ten candidates per iteration for up to 100 iterations, in total 1,000 evaluations. The goal in all cases was to reach the global minimum of the original function, equivalently the global maximum of its negated form. Each experiment was independently repeated twenty times to capture statistical variability. At every iteration, the framework recorded (i) the best-so-far objective value, (ii) the minimal feature-space distance between the proposed candidates and the known global optimum, and (iii) ensemble-level model diagnostics, including uncertainty, information gain, and inter-model consensus. These metrics jointly characterize both the external convergence dynamics and the internal evolution of information states, enabling a comprehensive assessment of how efficiently the algorithm approaches the optimum and how its inference structure adapts under different landscape complexities.

\begin{figure*}[h]
    \centering
    \includegraphics[width=\linewidth]{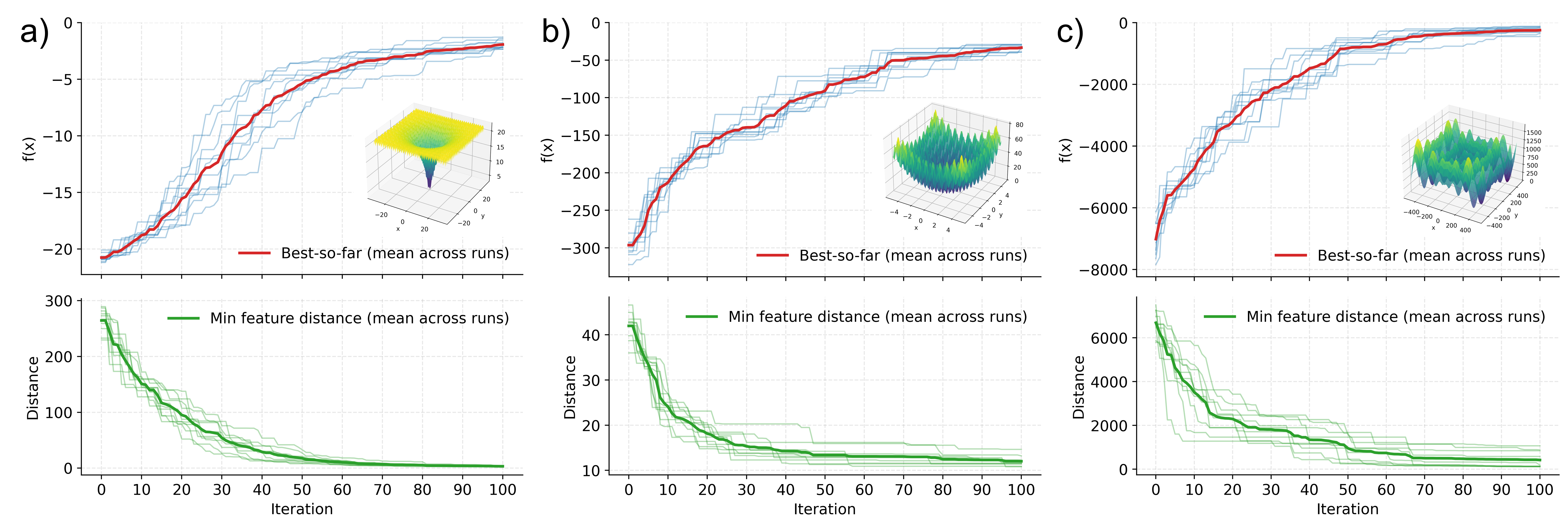}
    \caption{Convergence on 20-dimensional mathematical benchmarks of a) Ackley; b) Rastrigin; c) Schwefel function. 
    Upper panels: best-so-far objective (mean across runs; thin lines show individual runs), with the small panel in the middle is the visualization of the 2D examples for each mathematical function.
    Lower panels: minimum feature-space distance to the global optimum among proposed candidates (mean across runs; thin lines show individual runs). 
    Initialization uses 10 points; each iteration proposes a batch of 10 candidates; 
    results are averaged over 10 independent runs.}
    \label{fig:math_res}
\end{figure*}

Figure~\ref{fig:math_res} summarizes the convergence behavior of the framework across twenty independent runs for the three benchmark functions. In each panel, the upper plot shows the evolution of the best-so-far objective value, with the thick red line indicating the mean across runs and the thin blue lines representing individual trajectories; the lower plot reports the minimal feature-space distance between the best predicted point and the true global optimum, with the thick and light green lines denoting the mean and individual results, respectively. 

Across all functions, the framework exhibits consistent, monotonic improvement, demonstrating stable convergence under markedly different landscape complexities. For the Ackley function (cf. Figure~\ref{fig:math_res}.a), the best-so-far values rise smoothly toward the optimum and all individual runs follow very close trajectories with only minor deviations, indicating strong repeatability. The feature-distance curve decays almost exponentially, approaching zero within about 60 iterations, showing rapid localization within the central valley. For the Rastrigin function(cf. Figure~\ref{fig:math_res}.b), the trajectories progress in a stepwise pattern corresponding to successive barrier crossings; despite dense multimodality, all runs converge to near-optimal regions with small residual error, while the distance curves display a sharp early drop followed by gradual refinement, reflecting fast identification of global basins and focused exploitation. For the Schwefel function(cf. Figure~\ref{fig:math_res}.c), the trajectories vary more across runs due to its deceptive landscape, yet the mean trend steadily approaches the optimum, with roughly half of the runs reaching the global minimum and the rest stabilizing near high-quality local optima. The distance curves exhibit large early fluctuations—signatures of long-range exploratory jumps—that diminish once the global basin is reached, illustrating the algorithm’s ability to trigger high-gain transitions enabling escape from deceptive minima. Overall, the narrow spread of trajectories and the consistently decaying feature distances underscore the robustness and sample efficiency of the proposed information-fusion strategy, confirming its stable convergence across smooth, multimodal, and deceptive landscapes alike.

Having established the global behavior, we next examine representative single-run dynamics to expose how inference and sampling co-evolve under specific structural pressures as shown in Fig \ref{fig:math_analyze_result}. Each figure integrates (a) iteration-wise objective and feature-distance traces with cluster-wise bandwidth/quality evolution, (b) internal similarity, novelty versus history, and low-dimensional UMAP structure, and (c) model-level $R^2$–ELPD diagnostics and aggregated score ratios. Together these panes link information allocation, structural exploration–exploitation balance, and scoring-channel interactions within one trajectory, enabling a mechanistic account of why and how the method accelerates toward the global optimum.

\begin{center} 
    \includegraphics[width=\textwidth]{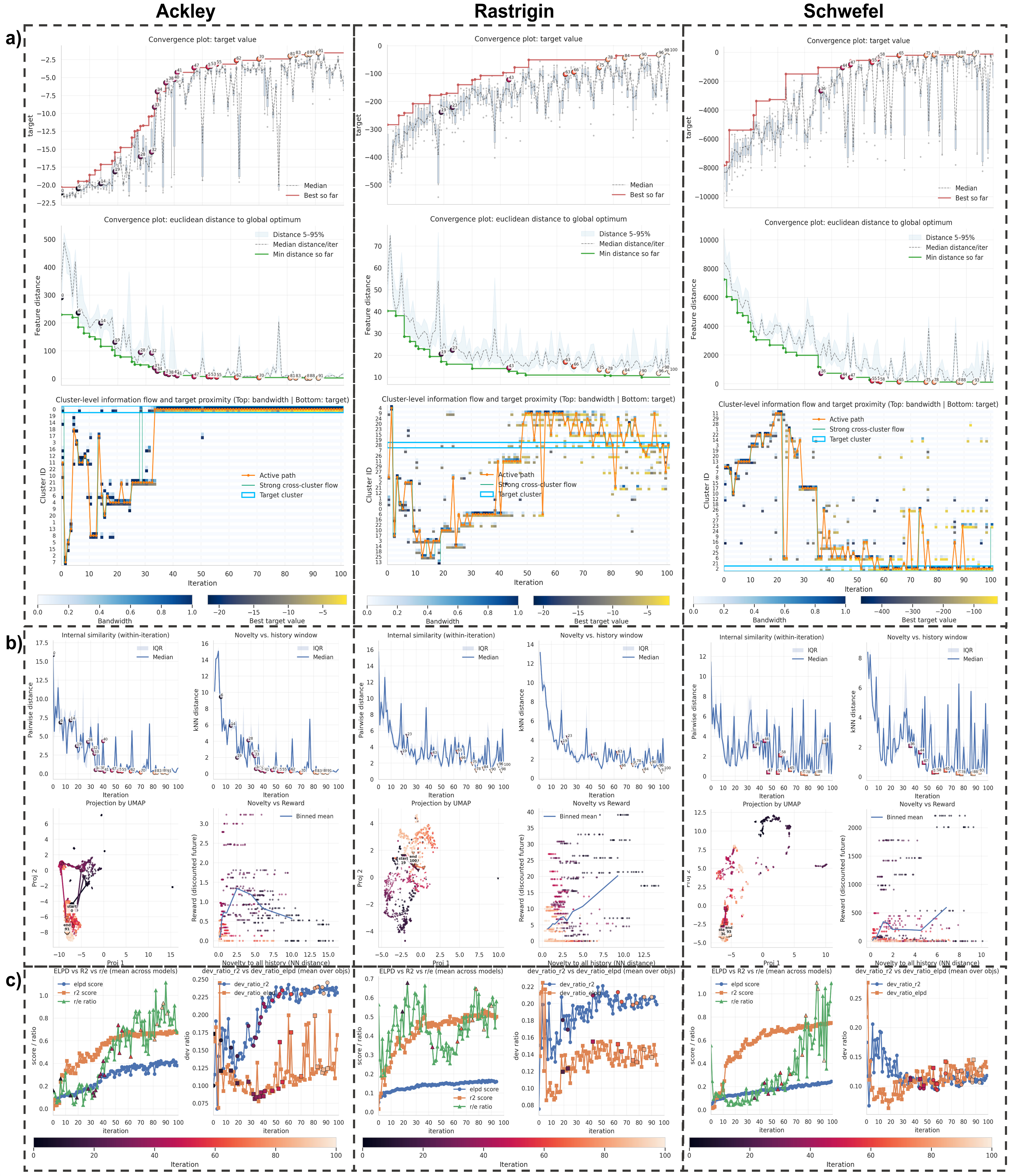}
\end{center}
\captionof{figure}{Composite visualization for Ackley test No. 3, Rastrigin test No. 18 and Schwefel test No. 5 of the multi-model optimization dynamics, data evolution, and performance analysis.
    a) Dynamics section: Iteration-wise convergence of the objective function and feature-space distance (top and middle), followed by the information bandwidth and quality evolution across clusters (bottom), showing how information flow concentrates and transitions among cluster manifolds during optimization; 
    b) Data analysis section: Internal similarity, novelty–history correlation, and the low-dimensional UMAP projection illustrate structural clustering, exploration–exploitation balance, and trajectory consistency over iterations; 
    c) Score section: Model-level evaluation metrics, including cross-model $R^2–ELPD$ consistency and aggregated score ratios, quantify the ensemble’s predictive stability and calibration through the optimization process.}
\label{fig:math_analyze_result}


Across all three benchmarks, the optimization dynamics exhibit a characteristic multi-phase pattern in which exploratory diffusion, cluster realignment, and concentrated exploitation alternate in a self-organized manner. Panel (a) shown in Fig \ref{fig:math_analyze_result} illustrates the optimization dynamics across 100 iterations (batch = 10). The top and middle plots track the best-so-far objective value and the Euclidean distance to the global optimum. Each boxplot and light blue region represent the distribution of the current batch, the dashed line the batch median, and the red/green curves the cumulative best of target value/feature distance. The numbers indicate the number of iterations by which a cluster containing the optimal point (the best cluster) first appears or updates the best target value within the cluster. The bottom plot reports the cluster-level allocation of sampling bandwidth (upper half of each cell) and the best target value within each cluster (lower half). Yellow lines trace the active path of progressive improvement, while green segments highlight strong inter-cluster transitions where the product of bandwidth transfer and quality gain exceed a defined threshold.  

In all cases, the best-so-far objective (top plot) follows a stepwise ascent, indicating discrete transition events rather than continuous improvement. The amplitude of these steps increases from Ackley to Rastrigin to Schwefel, consistent with the rising structural complexity of their landscapes. Each major jump in the objective is accompanied by a sharp drop in feature-space distance (middle plot) to the global optimum, followed by partial rebound and re-contraction, exhibiting an oscillatory rhythm of “temporary degradation → recovery → breakthrough.” For Rastrigin and Schwefel, the clusters containing the best samples typically emerge around a large transition, either immediately afterward or within the subsequent few iterations when sampling migrates from a neighboring high-gain cluster into the true optimum. This pattern indicates that informative regions are not discovered by gradual refinement but through occasional, high-variance transitions that open access to previously unvisited basins. In Ackley, by contrast, the globally optimal cluster already appears within the initial random batch and repeatedly re-emerges throughout the trajectory. This does not merely reflect the landscape’s simplicity but also reveals that the framework can infer global directional information even from extremely sparse data. Because the signal of the global trend dominates local fluctuations in Ackley, the ensemble can, through its consensus-weighted inference, identify a promising basin almost immediately and progressively contract its sampling bandwidth toward it.

The bottom plot shows that sampling bandwidth—the fraction of each batch allocated to different clusters—undergoes structured redistribution during optimization. In the early stage, bandwidth is broadly dispersed and frequent long-range transitions occur, maximizing directional information and preventing premature convergence. This behavior reflects the framework’s design principle: during exploration, the objective is not only to raise the target value rapidly but also to acquire global structural information. Consequently, the algorithm deliberately allocates sampling to clusters far from the current best region, enabling high-variance exploration across the landscape. At the same time, the evolution of bandwidth exhibits a directed tendency toward the global optimum. Because cluster indices in the plot are ordered by neighborhood proximity, it can be seen that most exploratory bandwidth during the early phase lies between the current dominant cluster and the true optimal cluster. As optimization proceeds, these intermediate clusters gradually accumulate higher-quality samples, and sampling progressively migrates toward the optimal cluster. In the intermediate stage, sampling concentrates around a few adjacent clusters that gather boundary evidence and refine the local direction toward the target basin. In the late stage, bandwidth collapses onto one or a few dominant clusters, yet a small residual allocation to peripheral clusters remains, ensuring minimal exploratory capability. This general three-stage pattern—broad exploration, neighborhood consolidation, and focused exploitation with residual probing—appears consistently across the three problems, but with landscape-specific signatures. Ackley exhibits numerous strong cross-cluster transitions early on and rapidly consolidates all bandwidth into a single dominant cluster. Rastrigin shows only one pronounced early jump and maintains a “one-main, several-auxiliary” bandwidth structure throughout, with medium-range transitions ensuring continuity across adjacent basins. Schwefel, the most deceptive landscape, displays strong cross-cluster jumps across all stages and retains low-level bandwidth in remote clusters even near convergence, maintaining persistent long-range exploration.

A vertical comparison across the top, middle, and bottom plots reveals that most significant objective improvements coincide with strong cross-cluster bandwidth transfers, indicating that performance jumps are accompanied by structural reallocation of sampling resources. This synchrony indicates that information flow reorganizes adaptively in response to newly acquired knowledge. When an exploratory move yields substantial improvement, subsequent iterations redirect sampling capacity toward the corresponding cluster, effectively reinforcing high-value information channels. From an information-theoretic perspective, the framework constructs an oriented path of entropy compression across the cluster manifold: exploratory diffusion generates informational diversity, while subsequent contraction focuses this flow toward regions of maximal target relevance. Collectively, these convergence behaviors reveal not only that the framework can identify and amplify informative directions under limited data, but also that it maintains a dynamic equilibrium between exploitation and exploration conditioned on problem complexity. In smooth landscapes (Ackley), the process behaves as a deterministic contraction toward a single basin; in multimodal or deceptive ones (Rastrigin, Schwefel), it reconfigures into a multi-path allocation scheme where auxiliary clusters act as risk-mitigating probes. Altogether, these coupled behaviors suggest that the framework’s convergence is driven by a closed feedback loop between external landscape signals and internal information reallocation, setting the stage for the data- and model-level adaptations discussed next.

Panel (b) shown in Fig \ref{fig:math_analyze_result} dissects the data-side evolution of the optimization process into four complementary views, arranged from top left to bottom right, that jointly reveal how structural diversity, novelty, and information gain evolve across iterations.
The first subplot quantifies internal similarity by summarizing pairwise feature distances within each batch, indicating when the search locally diversifies or contracts. The second tracks novelty with respect to history, computed as the average $k$-nearest-neighbor distance from current samples to the preceding three iterations, thereby measuring exploration intensity beyond recently visited regions. The third visualizes all samples via a two-dimensional UMAP projection fitted on adaptive cluster prototypes, showing how candidates migrate coherently along low-dimensional manifolds rather than through erratic jumps. The fourth relates each sample’s novelty to its discounted future reward, linking exploration amplitude to subsequent performance gains and highlighting the transition from exploratory to exploitative sampling regimes. Together these four views aim to expose the information geometry underlying each iteration, including how the algorithm expands its representational support, identifies new informative regions, and then contracts toward the target manifold as global structure becomes resolved.

Aligning panel (a) with the first two plots of panel (b) shows that major improvements are typically preceded by a 1–3-iteration “exploratory pulse”: simultaneous rises in intra-batch distance and novelty relative to recent history, matching the medium-/long-range cluster transfers in the bandwidth heatmaps. 
In general, these synchronized pulses indicate a brief decorrelation from recent samples that increases information gain, confirming that exploration is not random wandering but an orchestrated diversification aimed at maximizing directional evidence. In Ackley, after bandwidth has collapsed onto the dominant cluster, three late-stage improvement pulses persist without cross-cluster reallocations(iteration 50, 63, 78). Their intra-batch distance amplitudes remain similar while novelty with respect to recent history increases across pulses, indicating temporally-decorrelated, manifold-aligned forward steps within the same cluster. Consistent with this interpretation, the batchwise maximum feature distance to the global optimum remains roughly capped, not due to an explicit step-size limit, but because the scoring-and-selection pipeline implicitly recognizes a cluster/feasible boundary via prototype neighborhoods, uncertainty–consistency penalties, and diversity allocation), whereas the batchwise median distance expands approximately linearly across the three pulses which is evidence of a shell-like outward redistribution that allocates more samples to farther, yet still on-manifold, locations. The accompanying temporary degradation in target value (from -5 to -20) is expected in the wide, flat valley of high-dimensional Ackley: small angular deviations along the valley compound into large $l_2$ displacements, incurring value penalties while confirming the basin’s geometric extent and curvature before re-contraction. In contrast, Rastrigin exhibits broadened batches near the current region (novelty $<$ internal distance), consistent with parallel probing of many adjacent basins; Schwefel shows tight batches at new, remote locations (novelty $>$ internal distance), consistent with far relocations followed by focused local refinement.

The UMAP trajectories indicate that sampling generally progresses along manifold-aligned, band-like paths, albeit with intermittent gaps. For Rastrigin, the low-dimensional paths remain largely coherent with only minor discontinuities, and early long-range relocations are followed by short-range refinements near the target basin. Schwefel, while showing a similar early-stage pattern, retains occasional long-range excursions even in late iterations, reflecting the need to reassess deceptive basins and confirm global consistency. Ackley, by contrast, exhibits larger inter-segment gaps because its weak gradients and low signal-to-noise ratio make identifying the global direction inherently difficult. The early cross-cluster detours therefore serve as adaptive exploratory maneuvers to establish this direction before the search contracts back to the optimal region. All this behaviors highlights the framework’s capacity to adjust its exploration scale to the intrinsic difficulty of the landscape rather than engaging in blind diffusion. The overall piecewise coherence across functions corroborates panel (a)’s conclusion that the framework channels information flow along oriented manifold paths, resorting to intermittent long hops only when needed to validate geometry or escape ambiguity. The novelty–reward scatterplots (lower right) further quantify how exploration intensity couples to target improvement. For Ackley, rewards peak at moderate novelty ($\approx 2.5$), confirming that once the global direction is recognized, modest perturbations are most efficient. At higher novelty, the payoffs collapse to a two-point support at $\approx 0$ and $\approx 0.5$, reflecting within-cluster, shell-like confirmation probes in late iterations—temporally decorrelated forward steps that sometimes incur temporary penalties yet occasionally yield small gains while mapping basin geometry. For Rastrigin, rewards increase almost linearly with novelty, consistent with the value of high-variance exploration in a densely multimodal landscape where new basins carry meaningful information. For Schwefel, the curve is non-monotonic with two maxima: an early peak at moderate novelty ($\approx 1.5$) associated with reaching the distant global basin, and a later resurgence at high novelty ($\approx 7.5$) indicating renewed broad exploration required to escape deceptive attractors. These distinct patterns support the view again that the framework tunes its exploration bandwidth to each landscape’s signal-to-noise structure, contracting when global trends are clear and re-activating high-variance exploration when deception or ambiguity arises.

Building on the dynamics and data-structure patterns in Panels~(a)–(b), Panel~(c) examines how these exploration shifts are internalized by the ensemble through its out-of-bag (OOB) scores and fusion weights. The first subplot reports the ensemble-averaged OOB $R^2$ and $\mathrm{ELPD}$, together with their ratio $R^2/\mathrm{ELPD}$ used at selection time. In this framework, OOB $R^2$ acts as a high-frequency indicator of local fit: it reacts quickly when new high-value or tail samples allow tight interpolation in previously unexplored niches. OOB $\mathrm{ELPD}$ evolves more slowly, reflecting its role as a calibration-oriented measure of how well the ensemble’s predictive density aligns with the reweighted data distribution; it only adjusts once the $R^2$-driven acquisitions accumulate enough mass to reshape global trends. 

Across all three benchmarks, OOB $R^2$ and $\mathrm{ELPD}$ show distinct temporal signatures. $R^2$ rises rapidly and saturates once several extreme points near the target tail are correctly interpolated, whereas $\mathrm{ELPD}$ increases more nearly linearly as predictive mass is gradually reallocated toward the true distribution. Their ratio $R^2/\mathrm{ELPD}$ highlights landscape-specific behavior. For Ackley, the ratio initially oscillates around $0.1$ and, after roughly 30 iterations—coincident with the identification of the globally optimal cluster—rises and stabilizes around $0.8$–$1.2$. This indicates a progressive shift toward fine-grained interpolation once a reliable global basin has been found. For Rastrigin, this ratio exhibits a three-stage pattern: it starts closer to $0.7$ when the ensemble relies primarily on local discrimination inside dense multimodality, dips toward $0.3$ during a mid-run interval where a long-range relocation to the vicinity of the optimal cluster occurs (reflecting a temporary emphasis on global calibration to justify a basin switch), and finally recovers towards to $0.5$ in later iterations once the new basin is confirmed. Over the full run, it fluctuates in a band around $0.3$–$0.6$, consistent with the need to balance local precision and global trend maintenance. For Schwefel, the ratio remains near the $\sim 0.1$–$0.2$ level for an extended early phase, reflecting the priority of establishing a coherent global trend under deceptive structure, and only after about 60 iterations (shortly after a dominant cluster has stabilized) does it rise rapidly toward $0.6$–$1.0$, emphasizing more on $R^2$ guided exploitation. This pattern is consistent with a "trend-first, precision-later" strategy.

The second subplot summarizes the deviation ratios $\mathrm{dev\_ratio}_{R^2}$ and $\mathrm{dev\_ratio}_{\mathrm{ELPD}}$, which determine how the acquisition utility mixes standard variance-based Kalman fusion (KF) with the reverse variant $r\mathrm{KF}$. Each deviation ratio is computed based on normalized OOB scores as the product of channel-wise score mean and score variance, so high values indicate score channels that are both strong and heterogeneous across models. This corresponds to the intended role of $r\mathrm{KF}$: directing sampling toward regions that models regard as highly promising while remain uncertain. Such boundary points are expected to provide high mutual information with respect to the ensemble posterior and help reconcile model disagreement without overemphasizing clear outliers. Empirically, increases in the deviation ratios tend to precede or coincide with the novelty pulses observed in Panels~(a)–(b): when model disagreement within a score channel becomes larger, the resulting rise in $\mathrm{dev\_ratio}$ allocates more weight to the $r\mathrm{KF}$ component, which in turn promotes more exploratory, outward moves and larger intra-batch distances; as the search recontracts toward more familiar regions, both $\mathrm{dev\_ratio}$ and novelty decay.

After an initial transient with medium to large oscillations in the very early iterations (with the duration of this phase increasing from Schwefel to Rastrigin to Ackley), the three benchmarks exhibit distinct steady-state patterns. For Ackley, $\mathrm{dev\_ratio}_{R^2}$ grows and stabilizes near $\sim 0.23$, whereas $\mathrm{dev\_ratio}_{\mathrm{ELPD}}$ remains lower (around $\sim 0.10$) but still exhibits sizeable oscillations, indicating that reverse fusion is applied more strongly to the $R^2$ channel while residual calibration disagreement across models continues to drive non-negligible $r\mathrm{KF}$ activity in the $\mathrm{ELPD}$ channel. For Rastrigin, both deviation ratios rise with similar frequency but maintain a persistent offset (stabilizing around $\sim 0.225$ for $R^2$ and $\sim 0.125$ for $\mathrm{ELPD}$), indicative of a long-lived regime where pointwise discrimination and trend calibration are both actively involved in boundary sampling to cope with dense local oscillations. For Schwefel, $\mathrm{dev\_ratio}_{R^2}$ decreases while $\mathrm{dev\_ratio}_{\mathrm{ELPD}}$ increases in the early and mid stages; the two eventually meet near $\sim 0.125$ and subsequently oscillate in phase with comparable amplitude, reflecting the need to jointly refine trend and precision as deceptive basins are resolved.

Detailed per-model trajectories further clarify the division of labor behind these ensemble-level trends. The three low-information-capacity linear models (Lasso, Ridge, ElasticNet) rarely dominate the final performance but exhibit steadily increasing OOB $R^2$ in the early and mid stages, particularly on Schwefel, indicating that their globally smooth parameterization captures coarse large-scale trends without overreacting to local fluctuations. When their scores rise and their ensemble weights temporarily increase, the aggregated prediction tends to favor moves along these coarse directions, which in turn helps trigger long-range exploratory jumps such as those observed in Schwefel. These jumps do not always lead immediately to better objective values—indeed, they often coincide with transient deteriorations in both target value and distance to the optimum—but they provide valuable evidence for either confirming or falsifying candidate global directions at relatively low cost. In contrast, high-information-capacity models (MLP, LightGBM, and XGBoost) and medium-capacity models such as $k$-nearest neighbors are more sensitive to local structure and respond strongly when new high-target or boundary samples are acquired. Their OOB $R^2$ and $\mathrm{ELPD}$ show pronounced fluctuations aligned with the exploratory pulses, and their heterogeneous scores across models are a major contributor to the high-mean, high-variance configurations that $r\mathrm{KF}$ is designed to amplify. The resulting dynamics are therefore not driven by any single component but emerge from a coupled interaction: low-capacity models provide a conservative global trend compass, high-capacity models supply locally sensitive structure detectors, and the two score channels together regulate how much reverse fusion is used to probe contested boundary regions.

Taken together, the dynamics in Panel~(c) reveal that the framework integrates information across three interacting layers. At the data-fusion layer, the KF/$r\mathrm{KF}$ branch combines per-model predictive means and uncertainties into shared fused fields, allowing cross-model evidence to shape the local prediction landscape. At the model-weighting layer, both the KF branch and the UCB-based acquisition branch use OOB-derived model scores to modulate how strongly each surrogate contributes to prediction and utility signals. At the score-inference layer, the complementary $R^2$ and $\mathrm{ELPD}$ channels summarize two epistemic views (pointwise accuracy versus distribution-level calibration) and their deviation ratios regulate how strongly each view influences the reverse-fusion component. Together with the division of labor between low-capacity trend estimators and high-capacity local detectors, these three layers form a coherent feedback loop: differences in model behavior generate score-level reweighting, which reshapes fusion at both the predictive and acquisition levels, ultimately producing the adaptive exploration–exploitation patterns observed in Panels~(a) and~(b). 

\subsection{Performance across heterogeneous datasets}

We benchmarked the framework on fourteen materials datasets spanning nearly four orders of magnitude in size (from ${\sim}6\times10^2$ to ${\sim}4\times10^6$ samples) and more than two orders of magnitude in feature dimensionality (from 10 to over 1,000 features) \ref{tab:datasets_summary}. To provide a unified view of their structural and informational diversity, we quantified geometric and information-theoretic indicators for each dataset, including intrinsic dimensionality $d_{\mathrm{int}}$, effective resolution $R = N/d_{\mathrm{int}}$, and multi-objective conflict index $C_{\mathrm{obj}}$ (Table~\ref{tab:datasets_summary}). The resulting landscape spans extremely heterogeneous regimes: small, strongly underdetermined experimental sets such as \texttt{HEA\_hardness} and \texttt{curie\_temp} ($N<2{,}000$); mid-scale, moderate-complexity simulation sets such as \texttt{dielectric}, \texttt{3DSC\_TC\_nosoap}, and \texttt{MP} ($N\approx2\times10^3$--$5\times10^4$); and very large, high-dimensional datasets such as \texttt{QM9}, \texttt{alexandria\_icams}, and \texttt{omol}. ($N>5\times10^4$). 

\begin{table*}[h]
\centering
\small
\caption{Summary of materials datasets used for benchmarking, combining structural and informational complexity indicators. $N$: number of samples; $d$: feature dimensionality; $d_{\mathrm{int}}$: intrinsic dimensionality; $R = N/d_{\mathrm{int}}$: effective resolution; $C_{\mathrm{obj}}$: objective conflict index.}
\label{tab:datasets_summary}
\renewcommand{\arraystretch}{1.05}
\setlength{\tabcolsep}{2pt}
\begin{tabularx}{\textwidth}{l l r r r r r X}
\toprule
Dataset & Type & $N$ & $d$ & $d_{\mathrm{int}}$ & $R$ & $C_{\mathrm{obj}}$ & Target properties \\
\midrule
\texttt{HEA\_hardness}   & Exp-S & 635  & 145 & 11  & 58    & - & Hardness ($H$) \\
\texttt{curie\_temp\cite{long2021accelerating}}   & Exp-S & 1,749 & 140 & 22  & 80    & - & Curie temperature ($T_C$) \\
\texttt{Additive\_Mfg.}   & Exp-M & 2,167 & 10  & 6   & 361   & - & Relative density ($\rho_r$) \\
\texttt{steel\cite{dunn2020benchmarking}}      & Exp-M & 2,180 & 133 & 6   & 363   & 0.29 & 0.2\% proof Stress ($\sigma_{0.2}$), Ultimate tensile strength ($\sigma_{UTS}$) \\
\texttt{melting\_points\cite{bradley10jean}}      & Exp-M & 3,025 & 1,024 & 589 & 5  & - & Melting point ($T_m$) \\
\texttt{bandgap\cite{zhuo2018predicting}}         & Exp-M & 6,030 & 132 & 26  & 232   & - & Band gap ($E_g$) \\
\texttt{3DSC\_TC\_nosoap\cite{sommer20233dsc}}     & ExpSim-M & 5,773 & 148 & 25  & 231   & 1.02 & Critical temperature ($T_{critical}$), Energy above hull ($\Delta E_{hull}$), $E_g$, Magnetization ($M$)\\
\texttt{dielectric\cite{petousis2017high}}           & Sim-M & 4,762 & 271 & 31  & 154   & - & Refractive index ($n$) \\
\texttt{optimade\_2dmat\cite{zhou20192dmatpedia}}         & Sim-M & 6,351 & 274 & 34 & 187  & - & $E_g$\\
\texttt{MP\cite{jain2013commentary}}     & Sim-L & 52,613 & 271 & 33 & 1,594 & 0.98 & Formation energy ($E_f$), $M$, $E_g$ \\
\texttt{powerfactor\cite{lim2021extrapolative}}          & Sim-L & 94,250 & 28  & 13  & 7,250 & - & Power factor ($PF$) \\
\texttt{QM9\cite{ramakrishnan2014quantum}}                  & Sim-L & 133,139 & 1,025 & 820 & 162  & 0.60 & Heat capacity ($C_v$), Isotropic polarizability ($\alpha$), HOMO–LUMO gap ($\Delta E_{HL}$) \\
\texttt{alexandria\_icams\cite{schmidt2024improving}}          & Sim-L & 415,412 & 275 & 37 & 11,227 & 0.92 & $\Delta E_{hull}$, $M$, $E_g$ \\
\texttt{omol\cite{levine2025open}}                 & Sim-L & 3,986,754 & 128 & 15 & 265,784 & - & $\Delta E_{HL}$\\
\bottomrule
\end{tabularx}
\end{table*}

Closed-pool optimization on these materials datasets is substantially more challenging than on synthetic test functions. Beyond high feature dimensionality and large variations in sample size, several intertwined structural factors shape the effective difficulty: severe sparsity (with effective resolutions spanning $R{=}5$ to $2.7\times10^5$), intrinsic dimensionalities up to $d_{\mathrm{int}}>800$, heterogeneous local neighborhoods, and nontrivial candidate-pool topology. For multi-objective datasets, target conflicts further introduce competing descent directions and incompatible local basins. A complementary source of complexity arises from the entropy of the closed pool itself. Large datasets such as \texttt{omol} and \texttt{QM9} contain millions of unqueried entries forming a high-entropy combinatorial landscape in which informative submanifolds must be located, whereas small datasets such as \texttt{steel} or \texttt{HEA} suffer from sparse coverage and incomplete connectivity, making uncertainty dominated by missing information. Both extremes—oversized search domains and under-sampled manifolds—increase system entropy and challenge the optimizer to suppress irrelevant variation while concentrating exploration on target-relevant regions. These considerations motivate a systematic decomposition of dataset difficulty beyond feature count or dataset size alone.

\begin{figure*}[h]
    \centering
    \includegraphics[width=\linewidth]{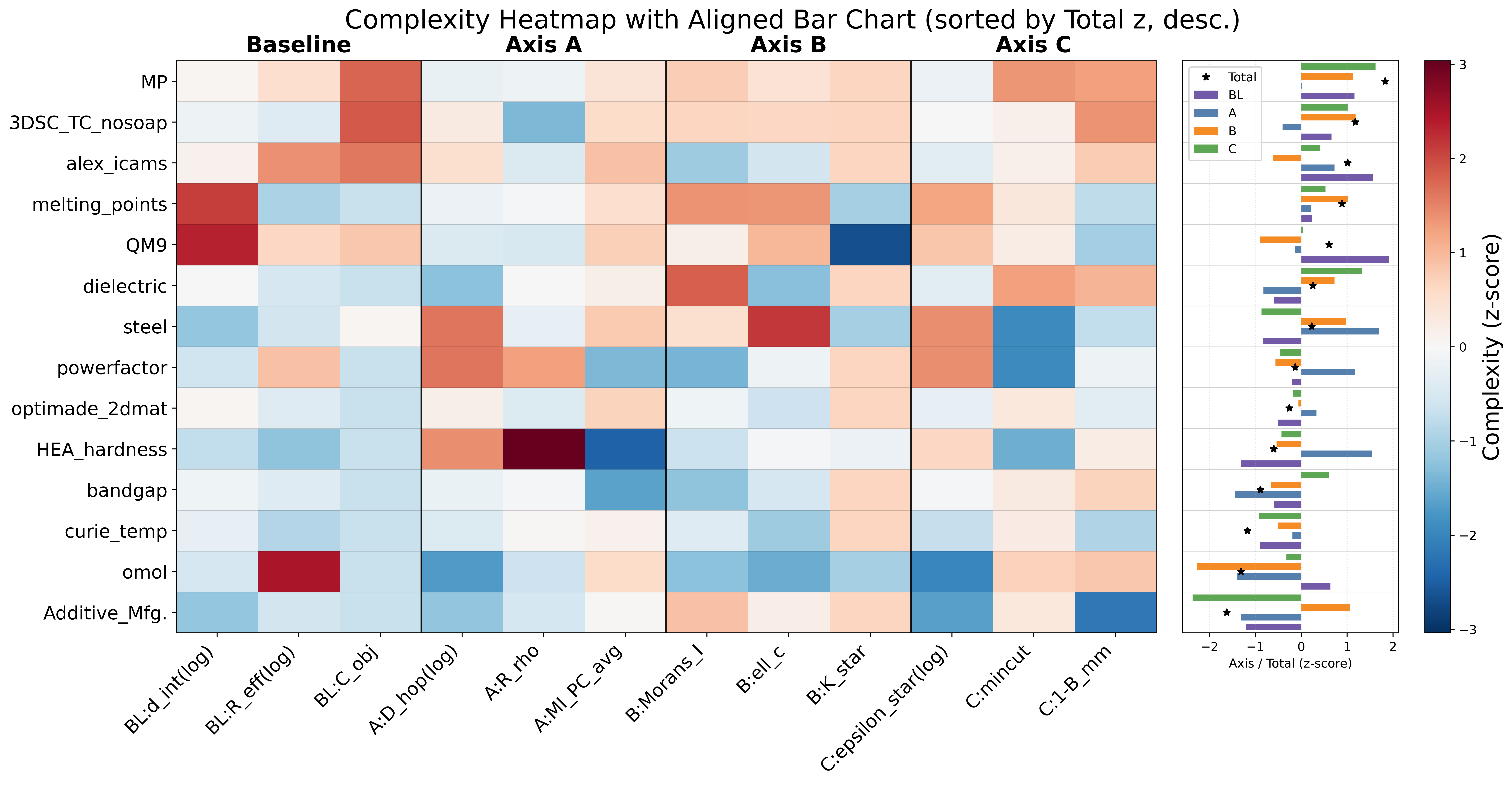}
    \caption{
    \textbf{Complexity heatmap across all datasets.}
    Each row corresponds to one dataset, sorted in descending order of the normalized total difficulty (hardness $z$-score). 
    Columns are grouped into four categories: Baseline metrics (\(d_{\mathrm{int}}\), effective resolution \(R_{\mathrm{eff}}=N/d_{\mathrm{int}}\), and objective conflict \(C_{\mathrm{obj}}\)); 
    Axis~A (information-scarcity: start--target graph distance $D_{\mathrm{hop}}$, target--neighborhood sparsity $R_{\rho}$, average mutual information between principal components and aggregation target $\mathrm{MI}_{\mathrm{PC,avg}}$); 
    Axis~B (heterogeneity and multimodality: Moran’s~\(I\), correlation length~\(\ell_c\), multimodality~\(K^*\)); 
    and Axis~C (deceptive and barrier-dominated: required jump~\(\epsilon^*\), min-cut capacity, bottleneck width~\(1-B_{\mathrm{mm}}\)).
    Color encodes the standardized hardness ($z$-score), with red indicating more difficult conditions and blue indicating easier ones; white denotes the average level.
    The right-aligned bar chart shows per-axis composite scores (Baseline, A–C) and overall total hardness, aligned with the heatmap rows.
    }
    \label{fig:data_complexity}
\end{figure*}

The multidimensional indicators therefore allow the datasets to be organized along three largely orthogonal difficulty axes as shown in heatmap \ref{fig:data_complexity}. The \textbf{A-axis} (information-scarcity) includes datasets such as \texttt{steel}, \texttt{powerfactor}, and \texttt{3DSC\_TC\_nosoap}, which exhibit large graph distances ($D_{\mathrm{hop}}\to\infty$), strong target sparsity ($R_\rho>2$), and weak feature–target coupling ($\mathrm{MI}^{\mathrm{PC}}_{\mathrm{avg}}<0.1$), requiring active information gathering under severe data limitations. The \textbf{B-axis} (heterogeneity and multimodality), represented by \texttt{3DSC\_TC\_nosoap}, \texttt{MP}, and \texttt{melting\_points}, is characterized by low spatial autocorrelation ($\mathrm{Moran}’s\,I<0.4$), short correlation lengths ($\ell_c<0.5$), and multiple structural modes ($K^\ast{=}5$), challenging the framework to fuse heterogeneous model and neighborhood information into coherent global trends. The \textbf{C-axis} (deceptive and barrier-dominated) includes \texttt{3DSC\_TC\_nosoap}, \texttt{MP}, and \texttt{dielectric}, which show narrow widest-path capacities ($B_{\mathrm{mm}}<0.1$), large cut values ($\mathrm{mincut}>30$), and long critical jump distances ($\epsilon^\star>80$), indicating topological bottlenecks where progress depends on infrequent, high-gain transitions. Together, these three axes span a broad spectrum of representative problem types—ranging from sparse and underdetermined systems to heterogeneous, multimodal, and barrier-dominated landscapes—providing a rigorous and diverse test bed for evaluating the framework under distinct informational regimes. Interestingly, the three difficulty axes identified for the materials datasets mirror the structural challenges encoded in the three mathematical benchmarks, respectively. In contrast to the mathematical functions each isolating one dominant failure mode, real materials datasets often present combinations of all three, making closed-pool optimization a compounded test of directional inference, heterogeneity resolution, and cross-basin exploration.

We next assess the proposed framework across these datasets to quantify its convergence efficiency and robustness under such varying levels of complexity. For every dataset, we initialize with 20 points whose property values do not exceed the dataset median. Each iteration proposes a batch of 10 candidates, if any of the top $min(10, 0.01N)$ candidate can be found within the budget limit (50 iterations, 500 samples), the run is considered successful and optimization stops; otherwise, the run is considered a failure. To quantify robustness under structural diversity, each dataset is repeated 10 times, sampling distinct initializations and acquisition trajectories. 

\begin{table*}[h]
\centering
\small
\caption{Performance summary across all materials datasets. For each dataset, we report the dominant difficulty axes, the overall success rate across 10 runs, the median number of iterations required to reach a top-performing candidate, and the 20–80\% interquartile range (IQR).}
\label{tab:performance}
\begin{tabularx}{\textwidth}{l c c c c}
\toprule
Dataset & Axis & Success rate [\%] & Median iters to success & IQR [20\%, 80\%] \\
\midrule
\texttt{HEA\_hardness}       & AC & 100 & 2 & [1, 3] \\
\texttt{curie\_temp}         & BC & 100 & 3 & [3, 6] \\
\texttt{Additive\_Mfg.}      & BC & 100 & 4 & [2, 11] \\
\texttt{steel}               & ABC & 100 & 2 & [1, 3] \\
\texttt{melting\_points}     & ABC & 100 & 7 & [2, 11] \\  
\texttt{bandgap}             & BC & 100 & 8 & [3, 12] \\
\texttt{3DSC\_TC\_nosoap}    & ABC & 100 & 16 & [5, 26] \\
\texttt{dielectric}          & BC & 100 & 6 & [4, 10] \\
\texttt{optimade\_2dmat}     & ABC & 100 & 9 & [4, 11] \\
\texttt{MP}                  & ABC & 100 & 21 & [13, 22] \\
\texttt{powerfactor}         & ABC & 100 & 15 & [5, 24] \\
\texttt{QM9}                 & ABC & 100 & 2 & [2, 4] \\
\texttt{alexandria\_icams}         & ABC & 100 & 15 & [5, 22] \\
\texttt{omol}                & AC & 90 & 4 & [1, 39] \\
\bottomrule
\end{tabularx}
\end{table*}

Table~\ref{tab:performance} shows that the framework maintains high convergence efficiency and robustness across all fourteen datasets despite their widely varying sizes and complexity levels.
Across the board, the success rate is 100\% for 13 out of 14 datasets and remains high (90\%) even for the most extreme case, \texttt{omol}. Median iterations to success stay well below the budget of 50 in every case: small datasets or the sets show lower difficulties on Axes~A/B/C (e.g.\ \texttt{HEA\_hardness}, \texttt{steel}, \texttt{bandgap}) are typically solved within 2–8 iterations, and even structurally heterogeneous, multi-objective datasets such as \texttt{3DSC\_TC\_nosoap}, \texttt{MP}, \texttt{powerfactor}, and \texttt{alexandria\_icams} converge in 15–21 iterations, far below the sampling budget. Notably, the ultra–high-dimensional \texttt{QM9} dataset ($d_{\mathrm{int}}>800$) reaches success in a median of only 2 iterations, indicating that high intrinsic dimensionality alone does not degrade performance when the underlying manifold is well populated.

The variation across datasets aligns more strongly with the composite difficulty axes than with $N$ or $d_{\mathrm{int}}$ alone. Datasets dominated by Axis~A (information scarcity) but with relatively simple topology (e.g.\ \texttt{HEA\_hardness}, \texttt{Additive\_Mfg.}) converge rapidly with narrow IQRs, while Axis~C–dominated or ABC–coupled cases with pronounced bottlenecks and multimodality (\texttt{3DSC\_TC\_nosoap}, \texttt{MP}, \texttt{powerfactor}, \texttt{alexandria\_icams}) show larger median iteration counts and wider IQRs. In ultra-large, high-resolution pools such as \texttt{omol}, the requirement to identify top-$10$ candidates among millions of entries effectively pushes the target into an extremely sharp tail of the distribution, so more information must be accumulated before a run achieves success, naturally broadening the effort distribution even though the median number of iterations remains low. Overall, these results indicate that the proposed framework reliably discovers top-performing candidates across a broad range of structural regimes, and that apparent increases in effort in highly heterogeneous or barrier-dominated datasets reflect a stricter success criterion rather than a breakdown of the underlying optimization mechanism.

To illustrate the framework’s behavior under representative informational regimes, we present detailed results for three materials datasets in the main text Fig \ref{fig:data_analyze}: \texttt{3DSC\_TC\_nosoap}, \texttt{alexandria\_icams}, and \texttt{QM9}. These datasets span the dominant baseline and difficulty axes identified in Fig.~\ref{fig:data_complexity}. Complete optimization results for the remaining eleven datasets—including ultra-large pools (\texttt{omol}, \texttt{MP}), sparse experimental sets (\texttt{HEA\_hardness}, \texttt{curie\_temp}, \texttt{bandgap}), and intermediate-scale or multimodal systems (\texttt{steel}, \texttt{melting\_points}, \texttt{powerfactor})—are provided in the Supplementary section.

\begin{center} 
    \includegraphics[width=\textwidth, height=1.2\textwidth]{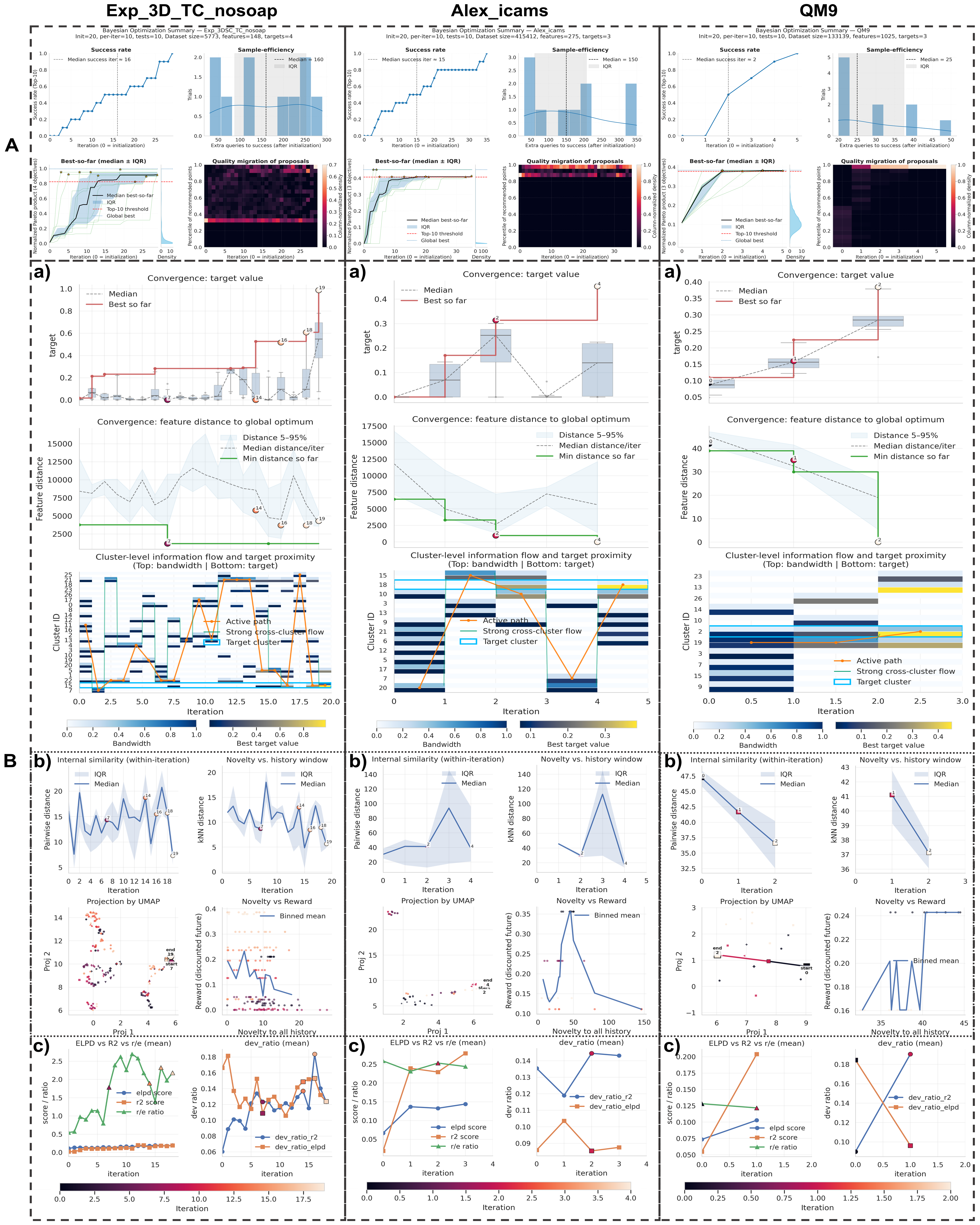}
\end{center}
\captionof{figure}{\textbf{Closed-pool optimization behavior on three representative materials datasets.}
    Columns correspond to \texttt{3DSC\_TC\_nosoap} (left), \texttt{alexandria\_icams} (middle), and \texttt{QM9} (right).  
    \textbf{A: Aggregate performance across 10 runs.}  
    Success-rate curves (top left), sample-efficiency distributions (top right), best-so-far convergence with property density distribution (bottom left) and proposal-quality migration (bottom right) summarize robustness and sample efficiency under distinct structural regimes.
    \textbf{B: Representative single-run analysis.}  
    Panels follow the same (a–b–c) decomposition used in the mathematical benchmarks: (a) dynamics and cluster-level bandwidth flow; (b) data-geometry evolution (similarity, novelty, and UMAP trajectories); and (c) model-level score adaptation ($R^2$, $\mathrm{ELPD}$, and deviation-ratio–driven $r$KF weighting).  
    Together these views show how the framework tailors its exploration–exploitation balance to dataset-specific difficulty: repeated cross-basin shifts in heterogeneous \texttt{3DSC\_TC\_nosoap}, gradual filtering in high-resolution \texttt{alexandria\_icams}, and rapid one-shot localization on the smooth, high-dimensional manifold of \texttt{QM9}.}
\label{fig:data_analyze}


Complementing the aggregate statistics in Fig.~\ref{fig:data_analyze}A, the bottom block (Fig.~\ref{fig:data_analyze}B) presents representative single-run trajectories for the same three datasets, using the same three-pane decomposition as in the mathematical benchmarks: dynamics and cluster-wise bandwidth (a), data-level structure and novelty (b), and model-level score evolution (c). 

For \texttt{3DSC\_TC\_nosoap}, the aggregate panel A shows a stepwise increase in success rate with a relatively broad and nearly uniform sample-efficiency distribution. The top-10 candidates are sparsely scattered in the long tail (roughly 0.8–1.0, cf. Best-so-far, Density plot), and the “quality migration’’ heatmap reveals that recommended samples span a broad range of the property distribution but gradually increase the ratio in the 0.9–1.0 band as iterations proceed. In the representative run, the dynamics panel exhibits pronounced stepwise gains in the best-so-far objective, whereas the distance to the global best decreases sharply only when the first high-quality cluster is discovered and then remains at a finite offset of order $4{\times}10^3$; this is consistent with a landscape containing several comparably attractive basins. Both median target and median distance undergo multiple sharp degradations followed by partial recoveries, each coinciding with strong cross-cluster bandwidth transfers in the lower panel: bandwidth is persistently distributed over several clusters, and only in the final iterations does it settle onto a small subset that dominates sampling. Data-level diagnostics show corresponding novelty pulses (simultaneous spikes in intra-batch distance and novelty relative to history) and a multi-lobed UMAP structure where the trajectory jumps between distinct manifolds. The novelty–reward relation rises over the course of the run, indicating that once sufficient structural information has been gathered, high-reward samples tend to arise at moderate to high novelty. On the model side, OOB $R^2$ remains slightly above $\mathrm{ELPD}$ but both absolute levels are modest, while the $R^2/\mathrm{ELPD}$ ratio is large, and the two dev\_ratios converge to similar levels and co-vary with novelty pulses. This suggests a search regime dominated by pointwise, $R^2$-driven exploration in which reverse fusion is repeatedly activated on both channels to probe high-disagreement boundary regions, which closely echos the Rastrigin/Schwefel-like behavior seen in the synthetic benchmarks.

For \texttt{alexandria\_icams}, which represents a very large but well-resolved closed pool, the aggregate statistics are still stair-like but markedly more concentrated. Most runs succeed within 0–200 additional queries, with only two late outliers approaching the 350-query limit. The distribution of top-10 samples lies in an extremely long tail (about 0.40–0.46) far away from main peak, yet the best-so-far curves approach this level within roughly five iterations, and even the slowest runs track very near the global best after about 25 iterations (cf. Best-so-far plot). The quality-migration heatmap shows that proposals already lie above the 80th percentile from the first iteration and quickly compress into the 0.85–1.0 range, with a progressive shift toward 0.95–1.0. In the single-run dynamics, best-so-far values improve smoothly with decreasing step size; after an initially diffuse phase, bandwidth rapidly locks onto a dominant cluster by the second iteration and finds the global best by the fourth, with only a few medium-range cluster jumps. In the data-level panel, intra-batch distances show moderate oscillations, but novelty against history decays rapidly once the high-quality region is located, and the UMAP trajectory forms a relative coherent band rather than hopping between disconnected components. The novelty–reward curve shows that moderate novelty yields the largest gains in early iterations, while the final iteration involves only small, locally refining moves—consistent with a setting where the main challenge is filtering a dense pool rather than crossing deep barriers. Correspondingly, OOB $R^2$ and $\mathrm{ELPD}$ rise in a more synchronized fashion, the $R^2/\mathrm{ELPD}$ ratio stays in a moderate range, and the two deviation ratios evolve in opposite directions with $\mathrm{dev\_ratio}\_{R^2}>\mathrm{dev\_ratio}\_{\mathrm{ELPD}}$, suggesting that the ensemble already captures the global distribution reasonably well and uses $R^2$ reverse fusion opportunistically to sharpen the decision boundary within the dominant basin rather than to initiate long-range relocations.

For \texttt{QM9}, difficulty is dominated by high intrinsic dimensionality and the need for long-range but geometry-aligned moves, while heterogeneity remains relatively low. The success curve rises almost linearly and saturates after only two effective iterations, with a sharply peaked sample-efficiency histogram; the best-so-far trajectories exhibit very narrow IQRs and reach the top-10 band (around 0.36–0.38) almost immediately. The proposal-quality heatmap shows that, after the first iteration, nearly all recommended molecules already lie in the top 5\% of the pool (0.95–1.0), and subsequent iterations merely refine within this range. In the representative run, both best-so-far value and distance improve nearly linearly over the few iterations before saturating, and cluster bandwidth collapses rapidly onto two neighboring clusters, each containing near-optimal candidates. Data-side diagnostics show a single, dominant novelty pulse in the first iteration, followed by approximately linear decreases in intra-batch distance and novelty; the UMAP trajectory traces a smooth, one-directional path along a well-populated manifold. The novelty–reward scatter contracts over time toward low-novelty, high-reward points, indicating that once the correct manifold region is identified, only small, geometry-aligned perturbations are needed to reach the optimum. On the score side, OOB $R^2$ and $\mathrm{ELPD}$ both rise within the first two iterations; by the second iteration $\mathrm{dev\_ratio}\_{R^2}$ exhibits a sharp increase while $\mathrm{dev\_ratio}\_{\mathrm{ELPD}}$ drops, consistent with a regime where calibration has already stabilized and reverse fusion is briefly emphasized in the $R^2$ channel to sharpen local geometry inside the correct basin.

Overall, these single-run analyses show that the same information-theoretic mechanisms of exploratory novelty pulses, bandwidth reallocation across clusters, and complementary $R^2$/$\mathrm{ELPD}$ channels modulated by deviation-ratio–controlled reverse fusion identified on synthetic benchmarks, reappear in realistic materials datasets but manifest differently depending on the dominant difficulty axis. Heterogeneous, multimodal systems such as \texttt{3DSC\_TC\_nosoap} rely on repeated, high-variance transitions and persistent boundary sampling; dense, high-resolution pools such as \texttt{alexandria\_icams} emphasize gradual filtering and local boundary sharpening; and high-dimensional yet well-populated manifolds such as \texttt{QM9} are solved by a single strong exploratory pulse followed by rapid contraction. This diversity of behaviors, together with the high success rates in Table~\ref{tab:performance}, supports the view again that the framework adapts its exploration–exploitation balance to the underlying informational regime rather than following a fixed sampling pattern. The closed-pool benchmark test on single target dataset of our method against other optimizers are shown in Supplementary.

In addition, we report two complementary sets of experiments in the Supplementary Information that further characterise the framework from a data perspective. The first group of tests probes the ability to identify candidates within a tightly constrained target interval, thereby assessing not only trend-following but also high-precision filtering. On the \texttt{alexandria\_icams} dataset with \(\sim 4\times 10^5\) entries, we consider a multi-objective interval-constrained task requiring simultaneous satisfaction of \(M \in [19.95, 20.05] ~\mu_\mathrm{B}\,\text{f.u.}^{-1}\), convex-hull distance \(< 0.03 ~\text{eV atom}^{-1}\), and band gap \(E_\mathrm{g} \in [1.45, 1.55] ~\text{eV}\). In two independent runs, the framework identified candidates meeting all three constraints within 27 and 35 iterations, with the found target values of $(M:19.993, d_{conv}: 0, E_\mathrm{g}: 1.509)$ and $(M:19.999, d_{conv}: 0.022, E_\mathrm{g}: 1.526)$, respectively, indicating that the proposed pipeline can support both coarse trajectory discovery and strict value-based screening when design targets impose narrow tolerances.

The second group of experiments focuses on the construction of physical descriptors and their interaction with the information pipeline. On the 3DSC dataset, the combined Magpie (148 dimensions) + SOAP (8715 dimensions) representation yields an 8863-dimensional feature space. When restricted to the Magpie subset alone, the framework already achieves competitive convergence; when using the full feature set with a strict cap of 30 iterations, 7 out of 10 independent runs still succeed, with only minor degradation compared to the low-dimensional setting. This suggests a degree of dimensional robustness: rather than relying on aggressive feature pruning, which risks discarding physically relevant but weakly correlated directions, we retain all candidate features and use the information budget and capacity control at the model level to filter noise. As a complementary perspective on representation quality, in the OMOL experiment we employ 128-dimensional descriptors extracted from a pre-trained universal molecular dynamics potential (UMA) as inputs for \(\sim 4\times 10^6\) molecules. Under tight budget constraints, the framework efficiently discovers high-performing candidates, indicating that high-quality learned representations are naturally compatible with the proposed information pipeline and offering a direct route to integrating large materials foundation models, via descriptor learning, into the physics axis \(\Phi_t\).

\section{Methods}
\label{sec:methods}

\subsection{Information-theoretic formulation: From function approximation to system entropy reduction}

Standard Bayesian optimization typically frames materials design as a function approximation problem: iteratively refining a surrogate model of the response surface \(f(x)\) to heuristicly estimate the optimum \(x^\star\)\cite{garnett2023bayesian}. This view implicitly assumes that the surrogate's inductive bias is well aligned with the underlying physics, an assumption that often breaks down in high-dimensional, data-scarce regimes. In such settings the primary difficulty is not to approximate \(f(x)\) everywhere, but to form a coherent belief about the latent optimality structure \(Z^\star\), representing the ground-truth configuration of the target-relevant manifold\cite{hennig2012entropy}.

Rather than treating the surrogate in isolation, we explicitly model the optimization process as the evolution of a tripartite information system:

\[
S_t = (D_t, \mathcal{M}_t, \Phi_t),
\]
comprising three orthogonal information axes (summarized in Algorithm~\ref{alg:entropy_bo}):
\begin{enumerate}
    \item Data axis (\(D_t\)): the set of grounded experimental or simulation observations accumulated up to iteration \(t\).
    \item Model axis (\(\mathcal{M}_t\)): a heterogeneous surrogate ensemble together with its uncertainty estimates and reliability weights.
    \item Physics axis (\(\Phi_t\)): structural constraints and manifold geometry of the candidate space, including external libraries, box constraints, and clustering-based geometry.
\end{enumerate}

Given this state, the residual uncertainty about the design outcome is captured by the conditional entropy \(H(Z^\star \mid S_t)\). Rather than maximizing an instantaneous utility, our objective is to design a policy \(\pi\) that minimizes this entropy under a strict evaluation budget. At each iteration we select the next query \(x_{t+1}\) from a feasible candidate set \(\mathcal{X}_t \subseteq \mathcal{X}\) so as to maximally reduce the expected uncertainty about \(Z^\star\):
\begin{equation}
    x_{t+1}^\star
    \;\in\;
    \arg\max_{x \in \mathcal{X}_t}
    I(Z^\star; y(x) \mid S_t),
    \label{eq:mi_system}
\end{equation}
where \(y(x)\) denotes the (possibly noisy and multi-objective) observation at \(x\), and \(I(\cdot;\cdot\mid\cdot)\) is the conditional mutual information\cite{hernandez2014predictive,wang2017max}. By symmetry,
\[
I(Z^\star; y(x)\mid S_t)
=
H(y(x)\mid S_t)
-
\mathbb{E}_{Z^\star \sim P(\cdot\mid S_t)}\!\big[ H(y(x)\mid S_t,Z^\star)\big],
\]
so that candidates are preferred when they exhibit high predictive uncertainty that can be resolved by information about the optimum, while task-irrelevant noise is downweighted.

Under standard conditional-independence assumptions, mutual information is a monotone submodular objective, and greedy selection of the point with the largest marginal information gain is near-optimal in idealized settings~\cite{nemhauser1978analysis,golovin2010adaptive,krause2008near}. Although we do not evaluate Eq.~(\ref{eq:mi_system}) in closed form, this result provides the conceptual baseline: our acquisition mechanism is designed as a tractable, ensemble-based proxy for \(I(Z^\star; y(x)\mid S_t)\), operating on a finite candidate set \(\mathcal{X}_t\).

\begin{algorithm}[t]
\caption{Entropy-aware multi-source Bayesian optimization}
\label{alg:entropy_bo}
\begin{algorithmic}

\State Given initial dataset $D_0$, domain specification (candidate pool $X_{\mathrm{pool}}$ or box constraints) and physical filters $\Phi_{\mathrm{phys}}$.
\State Initialize $t \gets 0$, $\mathcal{M}_0 \gets \emptyset$, $\Phi_0 \gets \Phi_{\mathrm{phys}}$, state $S_0 = (D_0,\mathcal{M}_0,\Phi_0)$.

\Repeat

  \State \textbf{Stage-0: Dimension-aware capacity control}
  \State Estimate effective dimension $d_{\mathrm{eff}}$ from current $(X_t,y_t)$.
  \State Compute information budget $N_{\mathrm{budget}}$.

  \State \textbf{Stage-1: Target-conditioned surrogate shaping}
  \State Fit heterogeneous ensemble $\mathcal{M}_t$ under $N_{\mathrm{budget}}$ using target-biased bootstraps.
  \State Compute OOB accuracy, calibration, stability and structural-trend reliabilities.

  \State \textbf{Stage-2: Structure-aware candidate construction}
  \If{$X_{\mathrm{pool}}$ is available}
    \State $X_{\mathrm{cand}} \gets X_{\mathrm{pool}}$
  \Else
    \State Generate $X_{\mathrm{cand}}$ via differential evolution and Sobol exploration.
  \EndIf
  \State Embed and cluster $X_{\mathrm{cand}}$; compute structural weights $\pi(x)$.
  \State Update structural/geometry component of $\Phi_t$ using the clustering of $X_{\mathrm{cand}}$.

  \State \textbf{Stage-3: Multi-source fusion and acquisition}
  \For{each $x \in X_{\mathrm{cand}}$}
    \State Compute family-level moments $(\mu_{a,t}(x),\sigma_{a,t}(x))$.
    \State Form per-model acquisition channel via reliability-weighted aggregation.
    \State Compute KF/rKF fusion channel using deviation ratios.
    \State Fuse channels via $\ell_2$ late fusion to obtain $A_t(x)$.
  \EndFor
  \State Rank candidates using $U_t(x) = \pi(x)\,A_t(x)$ (with MO-aware extension if $q>1$).
  \State Select batch $X_{\mathrm{next}} \subset X_{\mathrm{cand}}$ via diversity-aware ranking in $U_t(x)$.

  \State Evaluate oracle on $X_{\mathrm{next}}$ to obtain new pairs; update $D_{t+1}$.
  \State Set $S_{t+1} = (D_{t+1},\mathcal{M}_t,\Phi_t)$ and $t \gets 1 + t$.

\Until{$|D_t| \ge B$ \textbf{or} a desired target value is reached}

\State \Return final ensemble $\mathcal{M}_T$ and evaluation set $D_T$.

\end{algorithmic}
\end{algorithm}

\subsection{Constructing the tripartite information channel}

Directly optimizing Eq.~(\ref{eq:mi_system}) over a high-dimensional domain is infeasible. We therefore approximate it by shaping the evolution of \(S_t\) through a four-stage pipeline that aligns the three axes (Data, Model, Physics) to maximize information flow. Algorithm~\ref{alg:entropy_bo} summarizes the overall loop; here we outline the design principles of each stage. Detailed mathematical formulations and implementation choices are provided in the Supplementary Implementation Details.

\subsubsection*{Calibrating the model axis (Stages 0–1).}

The model axis \(\mathcal{M}_t\) is constructed as a particle approximation to the function-space posterior \(P(f \mid D_t)\). To avoid hallucinating structure in high-dimensional voids, we first perform dimension-aware capacity control (Stage~0). Motivated by manifold-based views of high-dimensional data~\cite{li2025basicsletdenoisinggenerative}, we estimate an effective intrinsic dimension \(d_{\mathrm{eff}}\) of the observed design–response pairs and map it to an information budget \(N_{\mathrm{budget}}\) that caps surrogate complexity and hyperparameter ranges\cite{levina2004maximum}. This ensures that model expressivity grows only in proportion to the estimated complexity of the target-relevant manifold.

Given this budget, Stage~1 performs target-conditioned surrogate shaping. We build a heterogeneous ensemble of linear, tree-based and neural surrogates trained under a target-biased bootstrap distribution derived from clustering in feature space\cite{lakshminarayanan2017simple}. High-performing samples within each structural cluster are upweighted in a “temperature-free” manner, effectively implementing importance sampling in data space. For each family, hyperparameters are optimized within the budget \(N_{\mathrm{budget}}\) and fixed; bootstrap training then yields out-of-bag (OOB) estimates of accuracy, calibration (e.g.\ \(R^2\), ELPD), structural trend fidelity, and stability. These OOB metrics are normalized into family-level reliability weights that later modulate both per-model acquisitions and fusion weights in Stage~3.

Stage~0 is re-evaluated at every iteration using the updated dataset \(D_t\), so that \(d_{\mathrm{eff}}\) and \(N_{\mathrm{budget}}\) adapt to the evolving resolution of the data manifold.

\subsubsection*{Constraining the physics axis (Stage 2).}

The physics axis \(\Phi_t\) defines the admissible search geometry. Rather than optimizing over the full ambient space, we construct a structure-aware candidate manifold \(\mathcal{X}_t\). In closed-pool settings, \(\mathcal{X}_t\) coincides with a fixed candidate library. In continuous or hybrid domains, Stage~2 generates a surrogate-implied candidate cloud by combining exploitation and exploration: for each surrogate family, differential evolution (DE) is run on a stochastic family-level objective constructed by randomly averaging a small subset of bootstrap members\cite{pant2020differential}; an additional DE run is performed on an ensemble-averaged objective; finally, a Sobol (or Latin hypercube) design is added as a geometry-aligned exploration baseline\cite{sobol2011construction}.

The union of these points is embedded into a low-dimensional space and clustered using an adaptive procedure that combines Gaussian mixture models, HDBSCAN, \(k\)-means, and Leiden partitions with multiple hyperparameter settings\cite{mcinnes2017hdbscan, traag2019louvain}. The selected partition yields cluster labels and local geometric descriptors (density, distance to cluster centres, boundary indicators), which are aggregated into a structural weight \(\pi(x)\in(0,1]\) for each candidate. Core points in well-understood regions receive slightly lower \(\pi(x)\), whereas boundary points, bridges between modes, and sparsely sampled tails are upweighted. The clustering information and structural weights update the physics component \(\Phi_t\), and \(\mathcal{X}_t = X_{\mathrm{cand}}\) serves as the finite support on which Eq.~(\ref{eq:mi_system}) is approximated.

\subsubsection*{Closing the loop via the data axis (Stage 3).}

Stage~3 fuses information across the three axes to compute acquisition scores on \(\mathcal{X}_t\). For each candidate \(x\), surrogate family \(a\), and target \(t\), we first compute family-level moments (means and variances) from the bootstrap members, and from these construct a dual-channel Signal-to-Noise Ratio that separates high-value peaks from broad improvement boundaries. Combined with structural trend metrics, these yield an exploration–exploitation ratio \(\gamma^{\mathrm{ee}}_{a,t}\in[0,1]\) that determines whether family \(a\) for target \(t\) is trusted primarily for exploitation (accuracy-driven, \(R^2\)) or for exploration (calibration-driven, ELPD).

A model acquisition channel is then formed by aggregating standard single-model acquisitions (e.g.\ mean–variance trade-offs) across families, weighted by their reliability and \(\gamma^{\mathrm{ee}}_{a,t}\). In parallel, a fusion channel is constructed via Kalman-like aggregation (KF) and its reverse counterpart (rKF) that amplify high-variance boundary regions. A deviation ratio \(\lambda^{\mathrm{dev}}_t\), computed from the mean and variance of normalized OOB scores, controls the balance between KF and rKF: high average performance with large spread triggers more weight on rKF, targeting regions where the ensemble is confident but internally diverse and thus information-rich.

For each target, the model acquisition channel and the KF/rKF fusion channel are combined via an \(\ell_2\) late fusion, producing a single target-wise score \(A_t(x)\). In the multi-objective case, the vector \(\mathbf{A}(x)=(A_1(x),\dots,A_q(x))^\top\) is further processed to account for data-driven inter-target dependencies: we estimate a residual covariance matrix from training residuals and use a Monte Carlo approximation to the expected hypervolume improvement (MC-EHVI) under this correlated posterior, yielding a multi-objective score \(U_{\mathrm{MO}}(x)\)\cite{daulton2020differentiable}.

Finally, the structural weight \(\pi(x)\) from Stage~2 is applied as a geometric prior,
\[
U(x) = \pi(x)\,S_{\mathrm{MO}}(x),
\]
and candidates are ranked by \(U(x)\). A cluster-aware tie-breaking rule ensures that the selected batch \(X_{\mathrm{next}}\) covers multiple structural modes. After evaluating \(X_{\mathrm{next}}\) with the expensive oracle, we update \(D_{t+1}\), recompute Stage~0 quantities, and advance the state to \(S_{t+1}\).

\subsection{Assumption analysis and pragmatic robustness}
\label{sec:assumption_robustness}

The particle-based approximations implicit in our framework rely on three weak assumptions: realizability (the true physics lies within the ensemble's hypothesis hull), coverage (the proposal \(\mathcal{X}_t\) overlaps with the true target region), and local smoothness (the response does not oscillate on scales finer than the spacing in $\mathcal{X}_t$). In complex material landscapes these assumptions may be locally violated, for example by feature blindness, “needle-in-a-haystack’’ optima, or high-frequency aliasing.

We mitigate these risks along two lines. Algorithmically, dimension-aware capacity control in Stage~0 and the heterogeneous ensemble in Stage~1 act at two complementary levels. Stage~0 imposes a global information budget tied to the effective data manifold, ensuring that the overall hypothesis space cannot grow arbitrarily more complex than what the observations can support. Within this constrained budget, Stage~1 distributes capacity across a spectrum of model families: low-capacity components act as deliberately under-parameterised surrogates, and high-capacity components focus on resolving finer structure. In parallel, Sobol and boundary injection in Stage~2 enforce a non-zero exploration probability and prevent the proposal distribution from collapsing, while the KF/rKF fusion in Stage~3 focuses sampling on informative disagreements rather than arbitrary high-variance artefacts.

More fundamentally, in realistic experimental workflows the primary success criterion is whether the procedure actually produces high-performing samples, rather than whether the surrogate offers a faithful reconstruction of the entire response surface. This pragmatic asymmetry gives us license to prioritise the data axis over the model axis: as long as the sequence of queried points (the projection of \(S_t\) onto \(D_t\) axis) progressively concentrates around satisfactory regions, the internal cognitive representation of the landscape may remain approximate. Accordingly, we can relax the learning objective from pointwise convergence \(|\hat{f}(x)-f(x)|\to 0\) to an order-preserving ranking on the candidate set,
\[
    f(x_A) > f(x_B) \;\Rightarrow\; \hat{f}(x_A) > \hat{f}(x_B),
\]
so that what matters is the stability of the induced ordering on \(\mathcal{X}_t\) rather than the exact numerical accuracy of \(\hat{f}\) at every point\cite{burges2010ranknet}.

Under this relaxed regime, the ensemble functions as a spectral filter bank whose components are controlled by the global budget from Stage~0: low-capacity families act as geometric low-pass filters that stabilize macroscopic trends\cite{li2025basicsletdenoisinggenerative} and maintain coverage, while high-capacity families act as high-pass filters that resolve local non-smoothness near the target tail\cite{rahaman2019spectral}. Even if realizability, coverage, or smoothness are locally violated, the system still converges toward high-utility regions in the sense of ranking, achieving rapid target discovery under stringent evaluation budgets.

\section{Conclusion}\label{sec:discussion}

This work proposes an information-theoretic reformulation of materials design: rather than viewing Bayesian optimisation in materials design as a global function-approximation problem, we treat it as the evolution of a tripartite system state \(S_t = (D_t,\mathcal{M}_t,\Phi_t)\) and aim to reduce the conditional entropy \(H(Z^\star \mid S_t)\) of the latent optimality structure \(Z^\star\). By explicitly separating the data, model, and physics axes and coupling them through a four-stage pipeline, the framework turns the intractable mutual-information objective \(I(Z^\star; y(x) \mid S_t)\) into a sequence of capacity-controlled, structure-aware, and consensus-driven decisions.

Conceptually, this perspective departs from classical BO in two ways. First, Stage~0 and Stage~1 implement dimension-aware capacity control and target-conditioned training, ensuring that surrogate expressivity grows only in proportion to an effective intrinsic dimension \(d_{\mathrm{eff}}\) inferred from the data. This prevents the ensemble from hallucinating structure in high-dimensional voids and focuses model capacity on resolving the target-aligned manifold. Second, Stage~2 and Stage~3 re-parameterise search in terms of a structured candidate manifold and a multi-channel acquisition rule. The candidate manifold constrains exploration to surrogate-supported trajectories and strategically retained stress points, while the acquisition combines per-model scores, Kalman-like consensus/disagreement (KF/rKF), and structural priors \(\pi(x)\) into a single proxy for mutual-information gain. Together, these mechanisms shift the optimisation objective from pointwise reconstruction of \(f(x)\) to maintaining a robust, order-preserving ranking on the candidate set, consistent with the practical priority of finding good materials over completely understanding the landscape.

Across high-dimensional synthetic benchmarks and realistic materials-design problems, this multi-source fusion strategy consistently exhibits three characteristic behaviours. First, it rapidly identifies and then tracks target-aligned trajectories, favouring queries that refine the geometry of promising channels rather than wasting measurements on flat, low-yield regions. Second, it remains robust under severe model mismatch: when individual surrogates misinterpret sharp physical transitions as noise or uncertainty, the ensemble-level reliability, SNR, and structural scores prevent the acquisition from being dominated by any single erroneous inductive bias. Third, by exploiting rKF-driven exploration of high-value disagreement regions, the method systematically allocates budget to stress tests of the ensemble’s current beliefs, enabling rapid entropy reduction even when the ground truth lies far outside the initial training prior.

At the same time, the theoretical justification of our approach rests on three weak yet nontrivial assumptions: realizability (the true physics lies within the ensemble hull), coverage (the candidate proposal overlaps with \(Z^\star\)), and local smoothness (discrete rankings preserve continuous trends). Many of our design choices can be read as targeted countermeasures: heterogeneous ensembles and capacity bottlenecks improve realizability; Sobol and boundary injection in Stage~2 safeguard coverage against mode collapse; restricting optimisation to an effective dimension \(d_{\mathrm{eff}}\) mitigates aliasing in high-dimensional ambient spaces. More fundamentally, by prioritising the data axis over the model axis and relaxing the goal from pointwise accuracy to ranking stability, the framework is engineered to degrade gracefully when these assumptions are only approximately satisfied. Low-capacity components act as geometric low-pass filters securing global coverage, while high-capacity components refine local structure near the target tail; the resulting spectral decomposition turns the ensemble into a filter bank that is robust to modelling imperfections.

Looking ahead, several directions appear particularly promising. On the modelling side, a tighter integration with learned descriptors and generative priors could further strengthen the coverage and realizability assumptions, for example via score-based models or diffusion-based candidate proposals. On the information-theoretic side, more principled approximations of mutual information and adaptive scheduling of the information budget across iterations may increase sample efficiency under extreme data scarcity. More broadly, our results suggest that robust progress in autonomous materials discovery may require shifting from value-centric Bayesian optimisation towards explicitly information-centric, trajectory-based design. By treating data, models, and physics as coupled channels in a unified entropy-reduction problem, the proposed framework offers a blueprint for future systems that can not only find better materials under tight budgets, but also reason about where information is missing and how to acquire it most effectively.

\backmatter


\bmhead{Acknowledgements}

The authors gratefully acknowledge the support by the Deutsche Forschungsgemeinschaft - Project-ID~405553726 - TRR 270. the computing time provided to them on the high-performance computer Lichtenberg at the NHR Centers NHR4CES at TU Darmstadt. This is funded by the Federal Ministry of Education and Research, and the state governments participating on the basis of the resolutions of the GWK for national high performance computing at universities (www.nhr-verein.de/unsere-partner).

\section*{Declarations}

\section*{Acknowledgements}

This work was supported by the Deutsche Forschungsgemeinschaft - Project-ID~405553726 - TRR 270. We also gratefully acknowledge the computing time provided to them on the high-performance computer Lichtenberg at the NHR Centers NHR4CES at TU Darmstadt. This is funded by the Federal Ministry of Education and Research, and the state governments participating on the basis of the resolutions of the GWK for national high-performance computing at universities (https://www.nhr-verein.de/unsere-partner).

\section*{Conflicts of interest}
There are no conflicts to declare.

\section*{Data and code availability}
All data needed to produce the work are available upon reasonable request from the corresponding author. All codes generated or used during the study are available in the github repository. 

\section*{Author contributions}
Yixuan Zhang: conceptualization, methodology, software, writing - original draft, validation, formal analysis, investigation, visualization, data curation;
Zhiyuan Li: conceptualization, methodology, software, formal analysis, validation, writing – review and editing;
Weijia He: conceptualization, methodology, software, validation, writing – review and editing;
Mian Dai: conceptualization, validation, writing – review and editing;
Chen Shen: writing – validation, review and editing;
Teng Long: conceptualization, methodology, supervision, writing – review and editing;
Hongbin Zhang: conceptualization, methodology, supervision, funding acquisition, project administration, writing – review and editing.

\noindent \textbf{Corresponding authors:} Yixuan Zhang, Teng Long and Hongbin Zhang (correspondence to: yixuan.zhang@tmm.tu-darmstadt.de, tenglong@sdu.edu.cn, hzhang@tmm.tu-darmstadt.de).\\[4pt]



\end{document}